\documentclass[1p,12pt]{elsarticle}

\usepackage{graphicx}
\usepackage{subfigure}
\usepackage{tabularx}
\usepackage{array}
\usepackage{bigstrut}
\usepackage{setspace}
\usepackage{ragged2e}
\usepackage{bm}
\usepackage{epstopdf}
	
\newcolumntype{P}[1]{>{\RaggedRight\hspace{0pt}}p{#1}}
\newcolumntype{C}[1]{>{\centering}m{#1}}

\usepackage[headsepline]{scrpage2}
\clearscrheadfoot
\ohead{Preprint of accepted manuscript for the Image and Vision Computing Journal (IMAVIS), doi:10.1016/j.imavis.2015.12.004}

\journal{Journal of Image and Vision Computing}

\begin{document}

\begin{frontmatter}



\title{Multimodal Classification of Events in Social Media}


\author{Matthias Zeppelzauer\corref{mycorrespondingauthor}}
\address{Media Computing Group, \\Institute of Creative Media Technologies, \\ St. P\"olten University of Applied Sciences \\ Matthias Corvinus-Strasse 15, 3100 St. P\"olten, Austria \\ m.zeppelzauer@fhstp.ac.at}


\author{Daniel Schopfhauser}
\address{Interactive Media Systems Group, \\Institute of Software Technology and Interactive Systems, \\Vienna University of Technology \\ Favoritenstrasse 9-11, 1040 Vienna, Austria \\ schopfhauser@ims.tuwien.ac.at}

\begin{abstract}
A large amount of social media hosted on platforms like Flickr and Instagram is related to social events. The task of social event classification refers to the distinction of event and non-event-related content as well as the classification of event types (e.g. sports events, concerts, etc.). In this paper, we provide an extensive study of textual, visual, as well as multimodal representations for social event classification. We investigate strengths and weaknesses of the modalities and study synergy effects between the modalities. Experimental results obtained with our multimodal representation outperform state-of-the-art methods and provide a new baseline for future research. 

\end{abstract}


\end{frontmatter}



\section{Introduction}
\label{sec:Intro}

Social media platforms host billions of images and videos uploaded by users and provide rich contextual data, such as tags, descriptions, locations, and ratings. This large amount of available data raises the demand for efficient indexing and retrieval methods. A tremendous amount of social media content is related to social events. A social event can be defined as being planned by people, attended by people and the event-related multimedia content is captured by people~\cite{reuter2013social}. The classification of social events is challenging because the event-related media exhibit heterogeneous content and metadata are often ambiguous or incomplete.

Indexing of social events comprises different tasks, such as linking media content belonging to a particular event (\emph{social event clustering} or \emph{social event detection})~\cite{petkos2012social} and summarizing the content of an event (\emph{event summarization})~\cite{delFabro2012summarization}. An important prerequisite for social event analysis is the distinction between event-related content and content that is not related to an event from a given stream of media. We refer to this task as \emph{social event relevance detection} or just \emph{event relevance detection}. After the selection of event-relevant content, a next task is the prediction of the event type. This task is referred to as \emph{social event type classification} or \emph{event type classification}.

\begin{figure} [t]
 \centering
  \subfigure[Fashion]{\label{sfig:intro1}
		\includegraphics[width=0.31\linewidth]{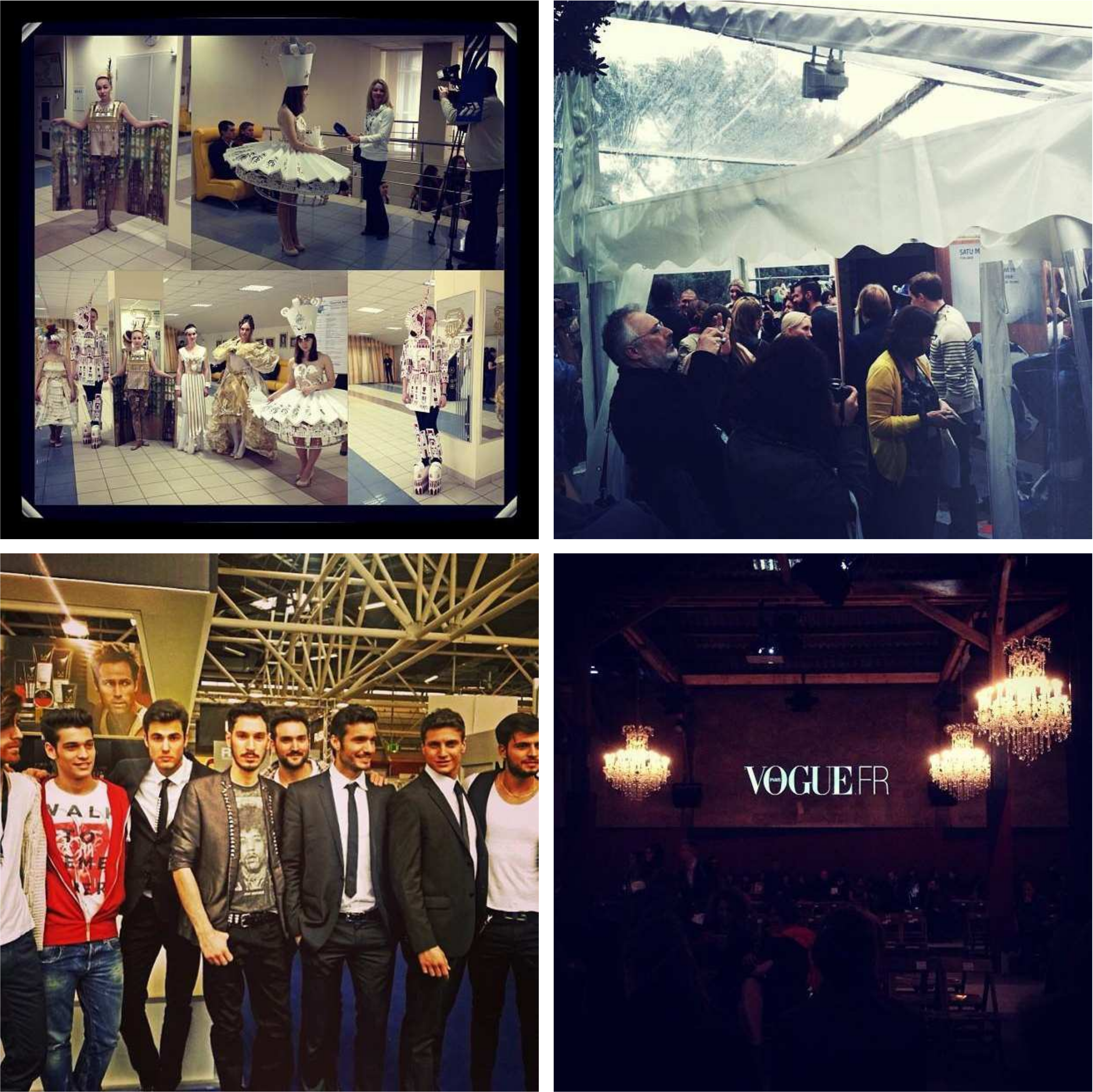}}
  \subfigure[Sports]{\label{sfig:intro2}
		\includegraphics[width=0.31\linewidth]{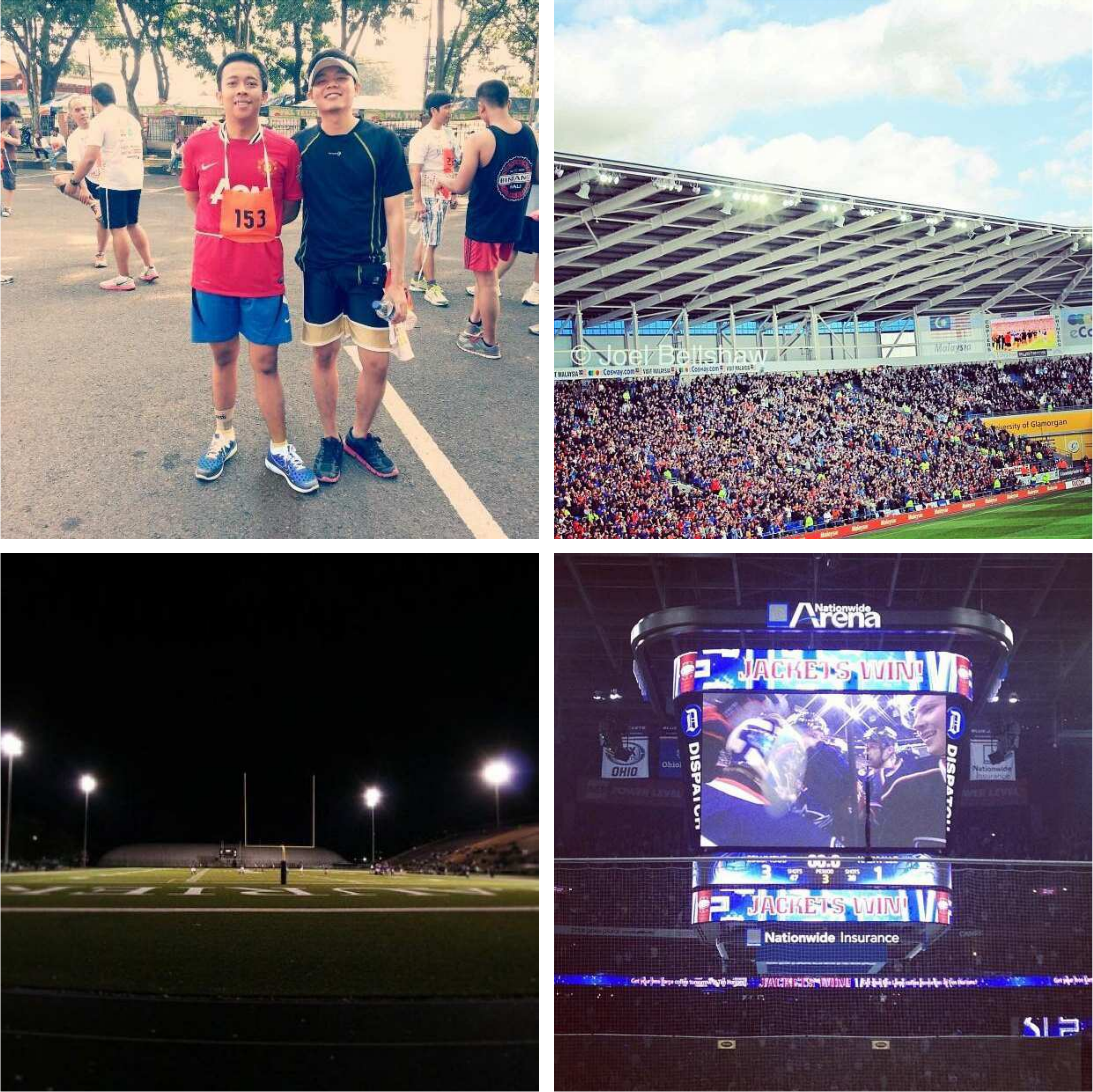}}
  \subfigure[Theater]{\label{sfig:intro3}
		\includegraphics[width=0.31\linewidth]{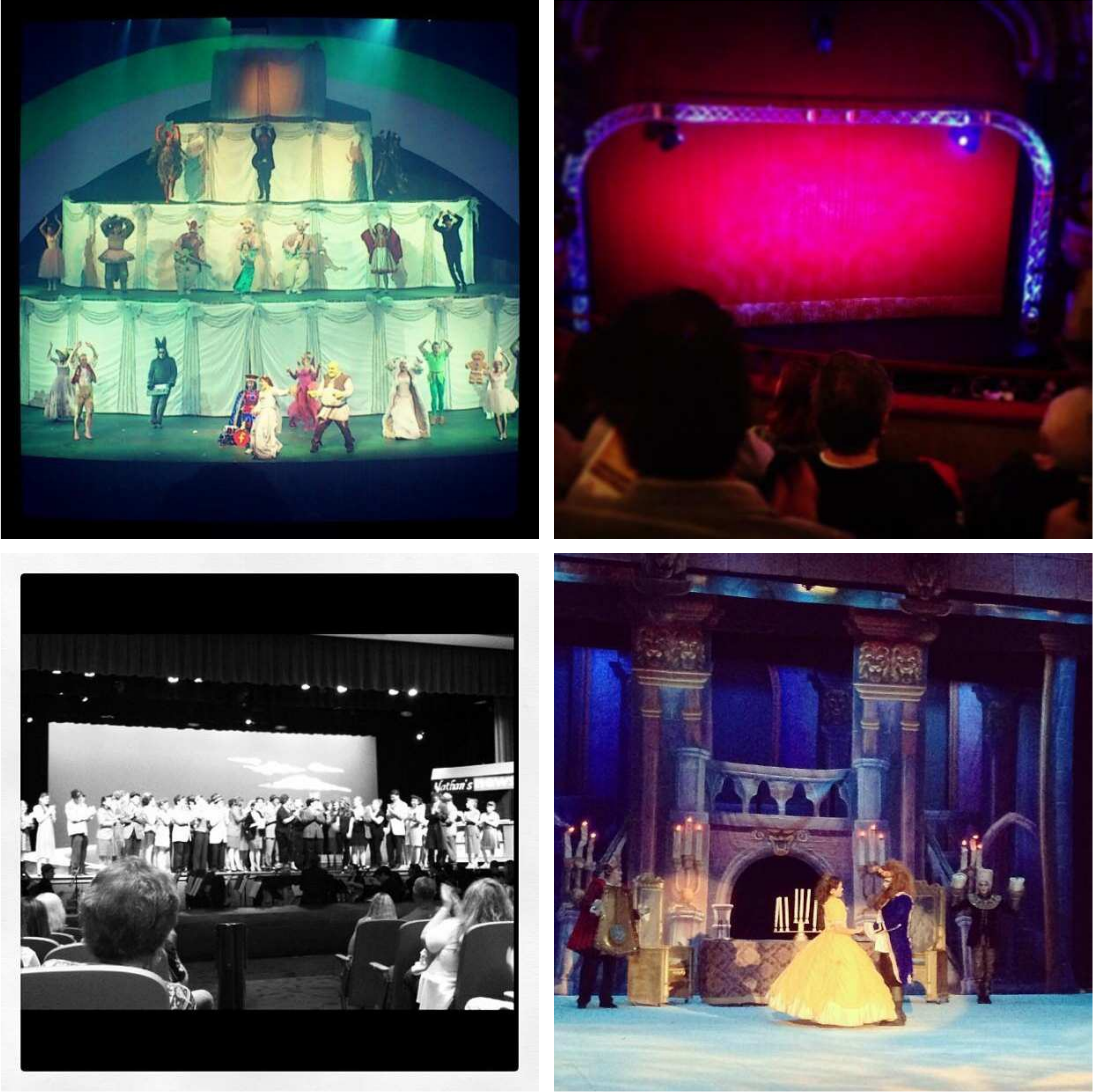}}
	\subfigure[Non-event]{\label{sfig:intro4}
		\includegraphics[width=0.965\linewidth]{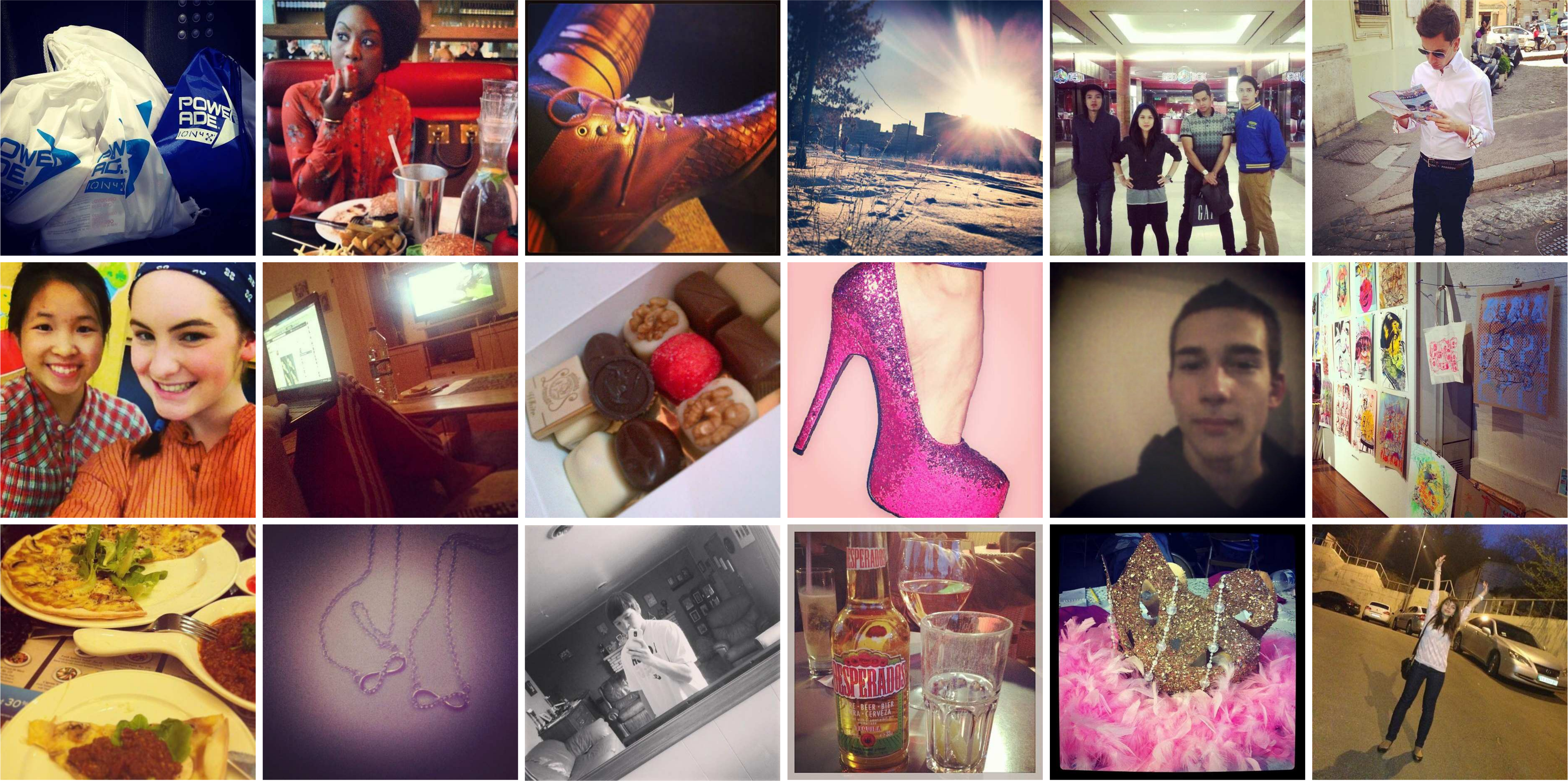}}
  \caption{Examples from three different event classes and of images that do not represent an event.}
\label{fig:imagedata}
\end{figure}

\begin{figure}[t]%
\centering
\includegraphics[width=0.8\linewidth]{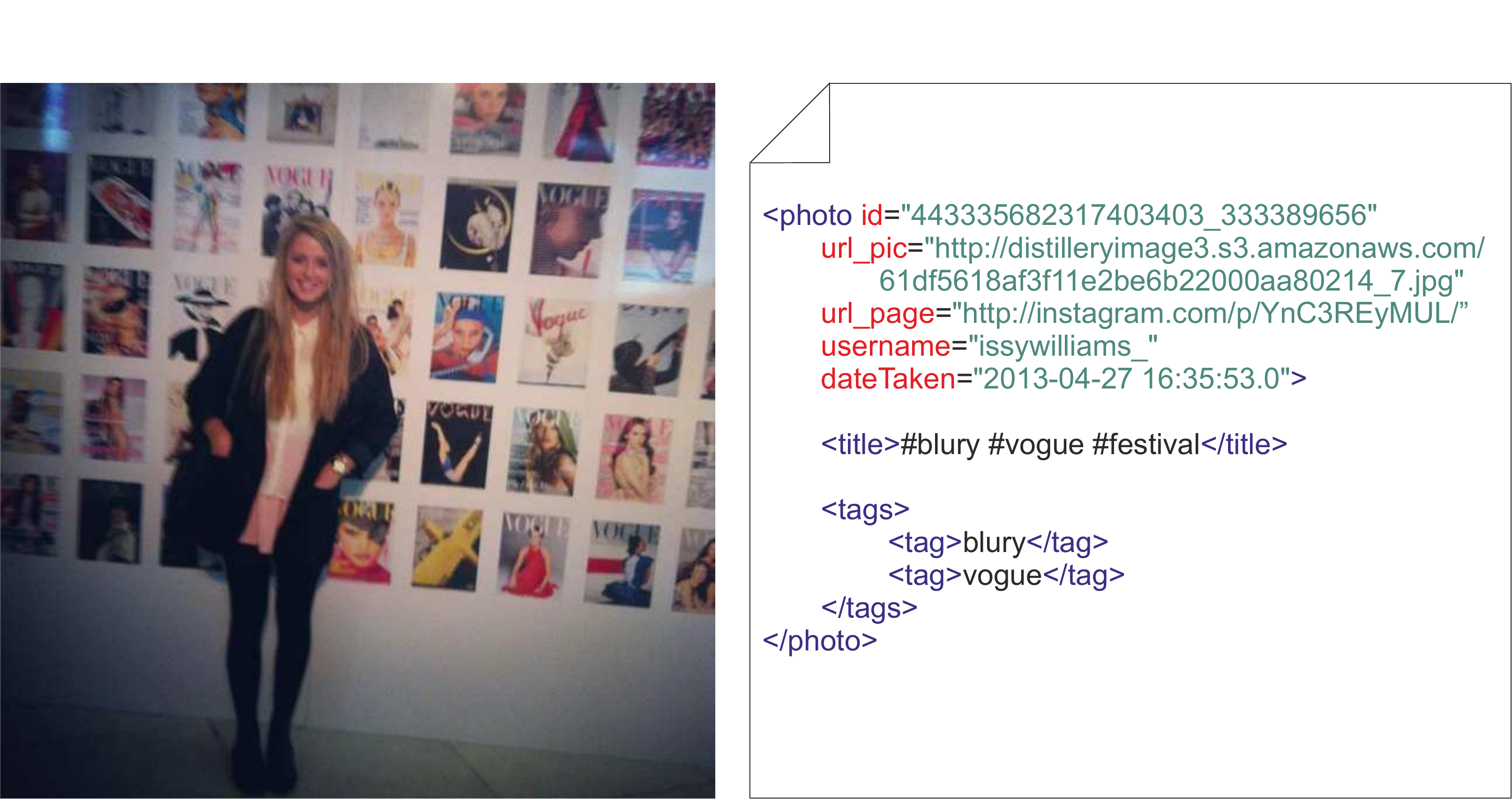}%
\caption{Ambiguities in the metadata of an image.}%
\label{fig:metadata}%
\end{figure}

The major challenges in the context of social event classification are (i) the high degree of heterogeneity of the visual media content showing social events, and (ii) the incompleteness and ambiguity of metadata generated by users. Figure~\ref{fig:imagedata} shows examples of event-related images as well as images without association to an event type (non-event images). We observe a strong visual heterogeneity inside the event classes. The non-event related images, however, are diverse and thus to find rules that separate them from event-related images is difficult. Figure~\ref{fig:metadata} illustrates an image with ambiguous metadata. The tag ``\#vogue" indicates a fashion-related event while ``\#festival" may also refer to a concert or musical event. The visual appearance of the related image resembles the appearance of the non-event images from Figure~\ref{sfig:intro4} rather than that of images from the ``fashion" class.

Recently, social event classification gained increased attention in the research community due to the availability of public  datasets and initiatives like the social event detection (SED) challenge of the Media Evaluation Benchmark~\cite{papadopoulos2011social,reuter2013social}. Partly motivated by the benchmark, numerous methods for social event classification have been introduced recently. Approaches employ either only textual metadata (contextual data)~\cite{ADMRG-SED2013, VIT-SED2013} or exclusively visual information~\cite{Imran2009, bossard2013event}. Only a few approaches combine contextual and visual information~\cite{brenner2014, Nguyen2014}. A comprehensive study of the multimodal nature of the task is currently missing and thus a focus of this work.

Our investigation comprises two tasks: (i) social event relevance detection and (ii) social event type classification. For both tasks, we evaluate the potential of the textual and visual modalities, investigate different multimodal representations and different information fusion schemes. We evaluate our method on the publicly available benchmark dataset from the SED 2013 challenge to enable direct comparison to related approaches~\cite{SED2013}. Our evaluation shows that multimodal processing bears a strong potential for both tasks. The proposed multimodal representation consisting of global and local visual descriptors as well as textual descriptors of different abstraction levels outperforms state-of-the-art approaches and provides a novel baseline for both tasks.

In Section~\ref{sec:RW} we review related work on social event classification. Section~\ref{sec:method} describes the proposed mono- and multimodal representations and different fusion strategies. The dataset and experimental setup are presented in Section~\ref{sec:exp}. We present detailed results in Section~\ref{sec:results} and draw conclusions in Section~\ref{sec:conclusions}.

\section{Related Work}
\label{sec:RW}

The detection and classification of event-related content has originally been proposed in the text retrieval domain. Early work in the field has been performed by Agarwal and Rambow~\cite{Agarwal2010}. The authors detect entities and their relations in text documents to infer events such as \emph{interaction event} or \emph{observation event}. 

With the increasing popularity of social media event detection and classification from images has become an attractive line of research. In~\cite{Imran2009} the authors present a purely image-based method to classify images into events like ``wedding" and ``road trip". The authors extract a Bag-of-Words (BoW) representation from dense SIFT and color features. Page rank is used for selecting the most important features and Support Vector Machines (SVM) finally predict the event type. The investigated dataset contains only event-related images. Hence, no event relevance detection is performed.  

Other works additionally exploit temporal constraints for event classification. Bossard et al.~\cite{bossard2013event} propose an approach for the classification of events from multiple images in personal photo collections. Again, only event-related images are considered. Similarly, the authors of~\cite{mattivi2011} integrate temporal constraints into the classification of events. 

The rich contextual metadata available through social media opens up new opportunities for event classification~\cite{dou2012event,Nurwidyantoro2012event}. A large benchmarking dataset comprising images together with contextual information is the Social Event Detection (SED) dataset from the Media Evaluation benchmark in 2013~\cite{SED2013}. It contains images together with metadata such as time, location, title and tags. The dataset as well as the SED challenge strongly promoted research in this area. Table~\ref{tab:RW} provides a systematic overview of recently developed approaches.

\begin{table}[t]%

\resizebox{\columnwidth}{!}{
\begin{tabular}{c}
    \includegraphics[width=1\textwidth]{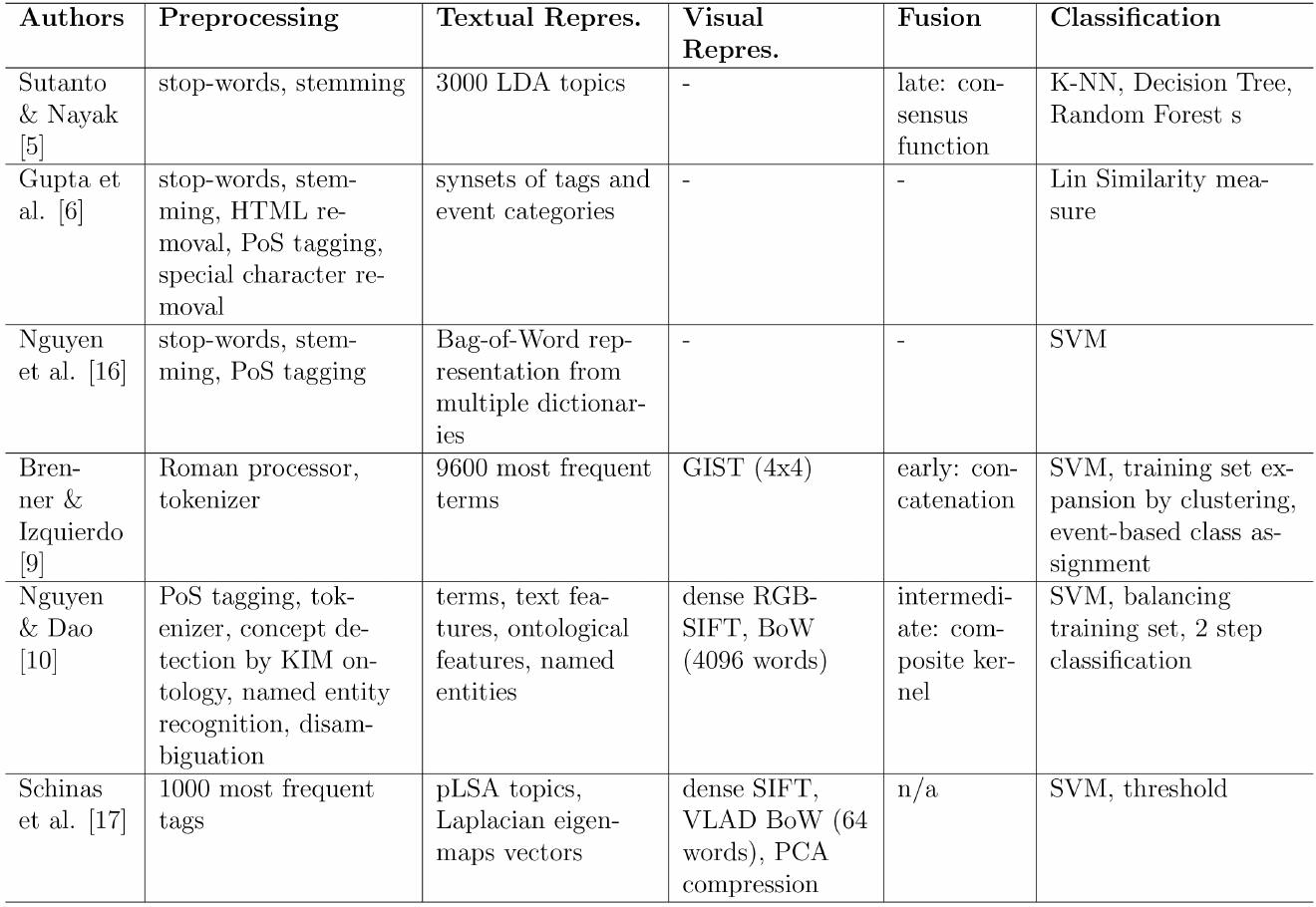}
\end{tabular}
}
\caption{State-of-the-art methods for social event classification and their building blocks.}
\label{tab:RW}
\end{table}

The first three methods in Table~\ref{tab:RW} are purely text-based. The remaining methods additionally incorporate visual information. The first step of all methods is a preprocessing of the textual data which includes stop-word removal, stemming, part-of-speech (PoS) tagging, and tokenization. Two approaches additionally use external information to extend the textual data (by WordNet and by ontologies)~\cite{VIT-SED2013,Nguyen2014}. 

The employed textual features are in most cases either the raw terms (e.g. the most frequent terms or tags) or topics extracted by latent semantic analysis (LSA) and Latent Dirichlet Allocation (LDA)~\cite{blei_etal:jmlr:2003}. Additional textual attributes employed are named entities and word types~\cite{Nguyen2014}. The multimodal approaches employ local features (dense SIFT) and build bag-of-word (BoW) representations from the descriptors~\cite{Nguyen2014,CERTH-SED2013}. The authors of~\cite{brenner2014} apply global features (GIST,~\cite{oliva2001modeling}) to capture the spatial composition of the images.

Event classification from social media is a multimodal task that incorporates text and images. A major challenge is the joint multimodal modeling of event classes. For this purpose, different fusion approaches have been proposed in literature. In early fusion textual and visual descriptors are appended to each other at feature level~\cite{brenner2014}. The joint modeling of event classes is thereby shifted to the classifier. In late fusion separate models are generated for image and text information which are then combined at decision level~\cite{ADMRG-SED2013}. Aside from early and late fusion, different strategies for intermediate fusion exist. Wang et al., for example, represent text and images by textual and visual words and fuse them by extracting joint latent topics using a multimodal extension of LDA~\cite{Wang2014Tomm,barnard2003matching}. The resulting topics capture mutual aspects of images and text. Another type of intermediate fusion is applied in~\cite{Nguyen2014}. The authors establish a multimodal feature space by combining the kernels (Gram matrices) of both modalities prior to classification.


For classification methods such as K-NN, decision trees, random forests, as well as support vector machines (SVM) are employed.~\cite{VIT-SED2013} directly apply similarity measurements to assign images to event categories instead of using a trained classifier.  

The overview of state-of-the-art methods shows that a wide range of different components (preprocessing steps, features, classification strategies) are applied. A comprehensive comparison of different techniques (and their combinations) is however missing as well as the investigation of the individual modalities' contributions. To fill this gap we provide a detailed investigation of different textual and visual representations for event classification and investigate the potentials of the individual modalities as well as that of their combination.

\section{Methodology}
\label{sec:method}

In this section we present techniques for preprocessing, media representation, and classification that we investigate in our study. We select techniques that have been successfully applied to social event classification and other social event mining tasks in the past as well as promising techniques that have not been applied for the task so far~\cite{jegou2010}.  In our study we combine the techniques to build mono- and multimodal approaches for event classification and evaluate their performance.

\subsection{Preprocessing}
\label{subsec:preproc}

Preprocessing focuses on the removal of unwanted information from the images' metadata (title and tags). We use all terms from title and all tags as input. A stop-word list is applied to remove words with low importance. Furthermore, we remove special characters, numbers, HTML tags, emoticons, punctuation, and terms with a word length below 4 characters.

\subsection{Textual representations}
\label{subsec:testrepr}

We employ two popular features to represent the textual information provided for each image: term frequency-inverse document frequency (TF-IDF) features~\cite{luhn1957statistical, zobel_moffat:sigir:1998, salton1988term, Aizawa200345}, as well as topics extracted from the preprocessed metadata. TF-IDF features represent the importance of words for documents (images) over the whole set of documents. In TF-IDF discriminatory terms are weighted stronger than terms that occur across many images (less expressive terms). TF-IDF is a popular and powerful but rather low-level feature that does not abstract from the available metadata.

Many different weighting schemes for TF-IDF exist. As shown in \cite{zobel_moffat:sigir:1998} the selection of a suitable scheme is a non-trivial task. We use the classic TF-IDF weighting scheme (``ntc" according to SMART notation \cite{Salton1971, manning2008}). A detailed investigation of different weighting schemes is out of scope of this investigation. The term frequency is the number of occurrences of a particular term in a document, i.e. ``natural" according to SMART notation. The document frequency is the number of documents the term appears in. We use the logged inverse document frequency, i.e. ``t" according to SMART notation. The resulting TF-IDF vectors are normalized to unit length by dividing each vector by its length (``cosine" according to SMART notation).
For TF-IDF computation, we employ the top $N$ terms (the $N$ most frequent terms in the collection) that remain after preprocessing. We compute TF-IDF representations of different dimensions for $N=$ \{500, 1000, 2500, 5000, 7000, 10000\}. Other selection strategies evaluated in preliminary experiments (Chi2-test, ANOVA, and an English dictionary) were rejected since they performed equally of slightly worse.

A more abstract representation is obtained by the extraction of \emph{topics}. For this purpose the preprocessed textual descriptions are assumed to be a mixture of latent topics. A robust and widely used method for discovering topics is Latent Dirichlet Allocation (LDA) \cite{blei_etal:jmlr:2003}. We employ the LDA implementation of the Mallet library \cite{McCallumMALLET} with Gibbs sampling. We generate multiple sets of topics with different dimensions, i.e. different numbers $T$ of topics: $T=$ \{50, 100, 250, 500\}. Each topic is associated with a number of words from the available metadata. The resulting feature vectors contain the topic probabilities of a given image over all $T$ topics. As the probability values for each image always sum up to 1 the feature vectors contain redundant information. An Isometric Log-Ratio Transformation is applied~\cite{egozcue2003} which maps the feature vectors to $T-1$ dimensional vectors with independent components. In our 	experiments, topics are extracted from the development dataset only, i.e. the test data is not incorporated in topic extraction.

\subsection{Visual representations}

We investigate two principally different and complementary types of visual features: global features and local features. For global description we select GIST features which represent the global spatial composition of an image~\cite{oliva2001modeling}. We expect that images from the same event type frequently show similar spatial layouts (e.g. the playing field in sports events). The GIST feature measures the orientation and energy of spatial frequencies across the image. The input image is first split into non-overlapping blocks. Next, for each block the frequency responses for a bank of Gabor filters with different orientations and scales are computed. The responses for each block and filter are aggregated and concatenated into a feature vector. The GIST feature vector represents information about the orientation and strength of edges in the different locations of an image and thereby gives an abstract global description of the scene.

The dimensionality of GIST strongly depends on the number of image blocks and the size of the filter-bank. We compute GIST for different numbers of blocks (1x1, 2x2, 4x4, 8x8, and 16x16) and a bank of 64 filters. Thus, for each image block 64 values are returned, which leads to a feature dimension of $16x16x64=16384$ for 16x16 blocks. As this high dimension leads to computational problems in classification, we reduce the dimensionality of the GIST features by PCA (PCA-GIST). After PCA we choose the minimum number of components necessary to reach a cumulative explained variance of 95\%. Feature vectors obtained from GIST or PCA-GIST are normalized to unit length prior to classification.

In contrast to global features, local features represent the fine structure of an image and neglect the spatial layout. We employ SIFT descriptors to describe the images and investigate sparse and dense sampling strategies~\cite{lowe2004}. To obtain a descriptor for classification we quantize the features and compute bag-of-words (BoW) representations. The codebooks necessary for the representations are created by choosing a class-stratified random subset of the development images. We employ $K$-Means clustering for codebook generation. Two different assignment strategies are used to create the BoW histograms: hard and soft assignment. 

In the hard assignment strategy each feature point of an image contributes only to one bin in the BoW histogram (that of the nearest codeword)~\cite{sivic2005}. We build such BoW histograms for sparse and dense SIFT points for different codebook sizes $K=$ \{500, 1000, 2500, 5000, 7000\}.

Additionally, we investigate a BoW representation with the soft assignment strategy VLAD (vector of linearly aggregated descriptors)~\cite{jegou2010}. In the VLAD representation the residuals between the input vectors and the nearest code words are encoded. The residual vectors for each code word are summed up and normalized. Finally, the normalized residual vectors for each code word are concatenated. In the VLAD encoding the codebook sizes are much smaller ($K=$ \{16, 24, 32\} in our experiments), leading to feature vectors of dimension 2048, 3072, 4096, respectively.

In a final step, we normalize the BoW histograms obtained from both assignment strategies by L2 normalization, i.e. the feature vectors are mapped to unit length. Normalization maps the feature values to similar value ranges and thus is an important prerequisite for the combination of different features during classification.

\subsection{Fusion and Classification}
\label{subsec:classification}

\begin{figure}[!ht]%
			\centering
				\includegraphics[width=0.93\linewidth]{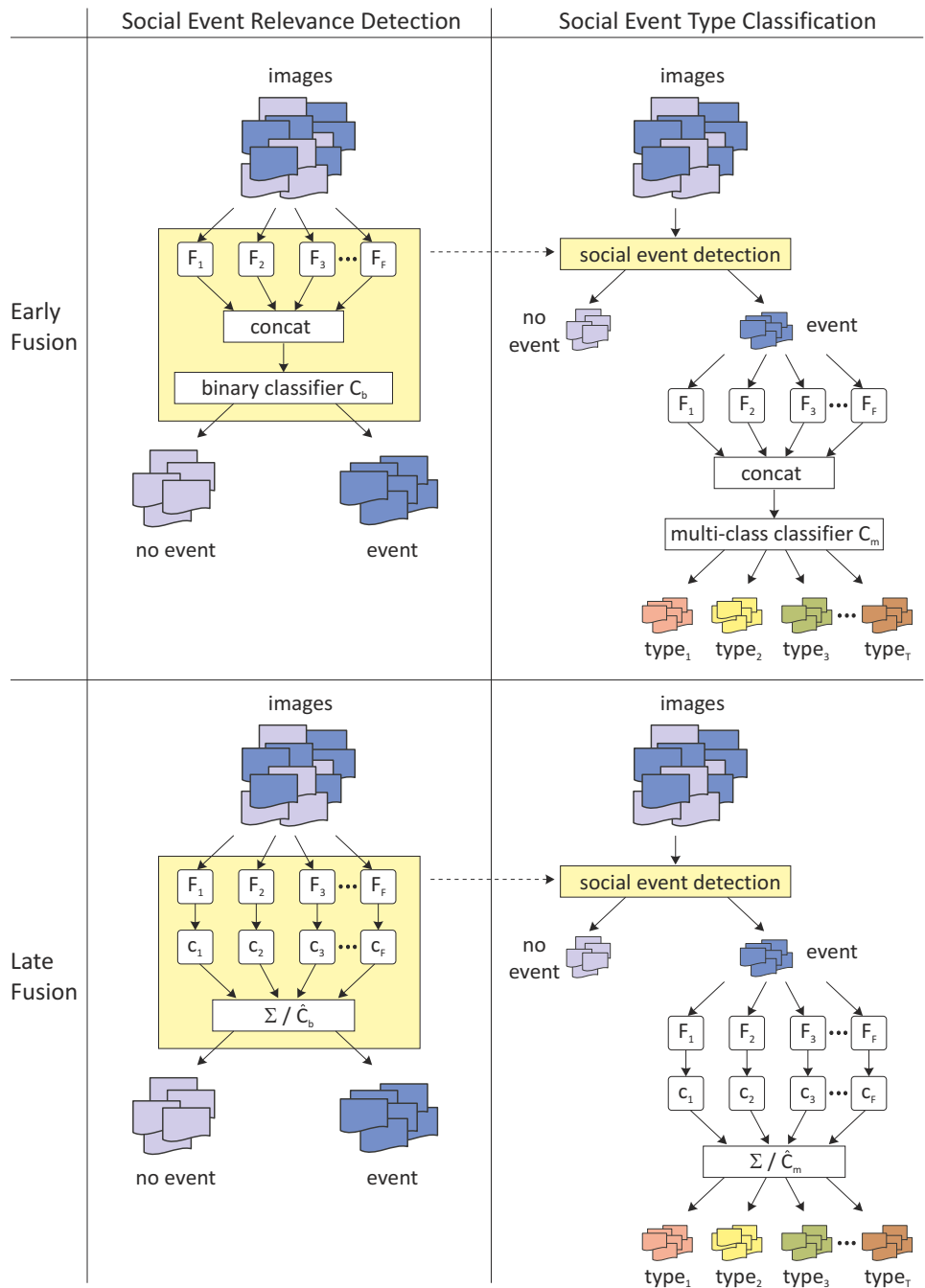}
			\caption{Classification schemes with early and late fusion for the investigated tasks. For both fusion schemes, event relevance detection is the basis for event type classification.}
			\label{fig:classificationStrategies}
	\end{figure}

As already mentioned in Section~\ref{sec:Intro}, we investigate two retrieval tasks: social event relevance detection and social event type classification. For both tasks we investigate classification strategies with early and late fusion. Figure \ref{fig:classificationStrategies} illustrates the different processing schemes.

For social event relevance detection with \emph{early fusion} we train a binary classifier $C_b$ from a set of concatenated input features $F_1, F_2,\ldots,F_F$ to separate event-related from non-event related images. For social event type classification we employ the trained classifier from event relevance detection to first identify event related images. Next, features for the remaining images are concatenated and a multi-class classifier $C_m$ is trained to distinguish the different event classes. The idea behind this hierarchical approach is to reject most non-event-related images in the first stage so that they do not interfere with the subsequent classification of event types in the second stage. This strategy allows the classifier $C_m$ to better adapt to the subtle differences between the event types.

Late fusion follows a similar scheme as early fusion. The major differences are that we train separate classifiers $c_1, c_2,\ldots,c_F$ for the input features and that each classifier outputs probabilities for the respective classes instead of predicted labels. We investigate two strategies to fuse the classifier's outputs: \emph{additive late fusion} and and \emph{hierarchical late fusion}. In additive late fusion the probabilities of all classifiers for an image are summed up (indicated by $\Sigma$ in Figure \ref{fig:classificationStrategies}) and the class with the highest accumulated probability is predicted. In hierarchical late fusion a separate classifier $\hat{C}$ is trained from the output probabilities of the lower-level classifiers $c_1, c_2,\ldots,c_F$ to generate a final class prediction.

\section{Evaluation}
\label{sec:exp}

In the following we introduce the employed dataset, the performance measures for both investigated tasks, specify the experimental setup, and state the major research questions behind our evaluation.

\subsection{Dataset}
To enable an objective comparison to a large set of existing state-of-the-art methods, we employ the publicly available benchmark dataset\footnote{Dataset available from: http://mklab.iti.gr/project/social-event-detection-2013-sed-2013-dataset} of the SED task from 2013 (challenge 2)~\cite{SED2013}. The dataset\footnote{Note that this dataset is different from the widely used dataset of SED challenge 1 for social event clustering} contains a total of 57165 images from Instagram with contextual metadata. Metadata consists of a title, a number of tags, the name of the uploading user, date and time of capturing, and partly geographic coordinates. 27.9\% of all images have geo information, 93.4\% have a title and 99.5\% have at least one tag. The vocabulary of the tags is uncontrolled and thus completely user defined. 

The dataset contains images from eight event classes and an additional (much larger) set of non-event-related images, see Table~\ref{tab:dataset}. The reason for the much larger non-event class is that the dataset creators observed that in practice only 1-2\% of images collected from a random stream are in fact related to events \cite{Petkos2014}. The imbalanced class cardinalities should reflect this asymmetry. The ground-truth has been generated by multiple human annotators~\cite{SED2013}. Borderline cases occurring during annotation were removed. 

\begin{table}[t]%

\centering
\resizebox{\columnwidth}{!}{
\begin{tabular}{c}
    
		\includegraphics[width=1\textwidth]{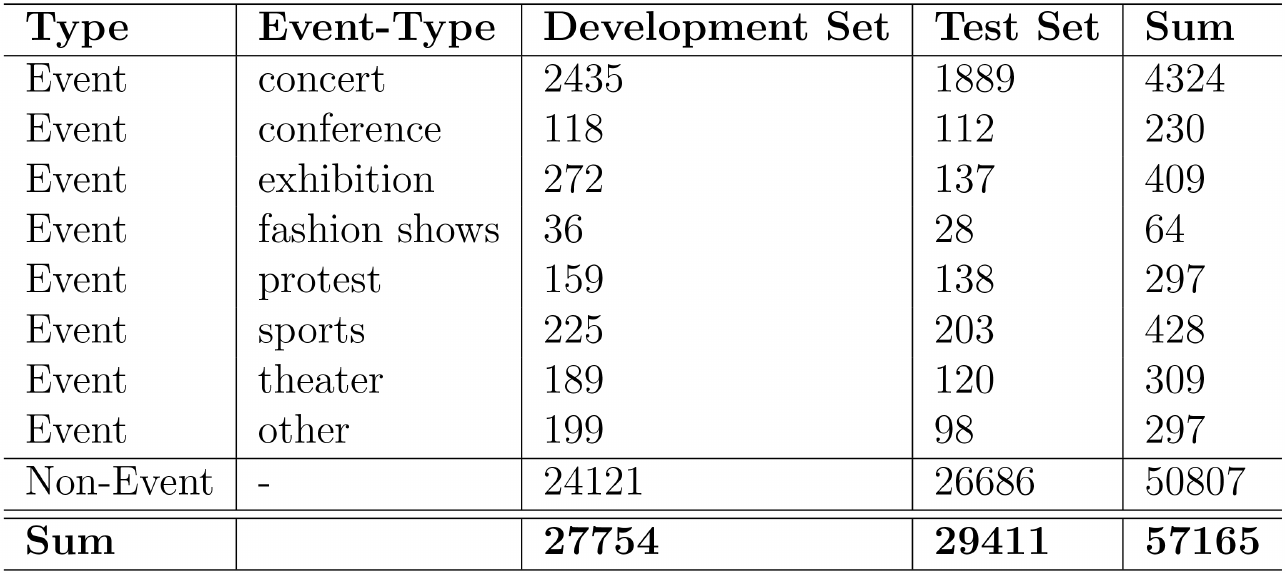}
\end{tabular}
}
 
\caption{The composition of the SED 2013 benchmark dataset.}
\label{tab:dataset}

\end{table}

\subsection{Experimental setup}

The focus of our evaluation lies on the investigation of different modalities and content representations for social event classification. Firstly, we study classification based on contextual information only and evaluate the boundaries of a purely textual approach. The best text-based approach serves as a baseline for all remaining experiments. Secondly, we investigate the suitability of the visual modality and evaluate different purely content-based representations. Thirdly, we add visual information to the (purely metadata-based) baseline approach and investigate the effects on performance. All investigations are performed for the two tasks (event relevance detection and event type classification) separately. 

For each evaluated representation we vary the most influential parameters (e.g. the number of clusters in BoW representations, the number of blocks for GIST, and the number of terms for TF-IDF) to investigate the sensitivity of each feature. 

For all experiments we employ the predefined development and test sets as defined in by the SED challenge. To estimate optimal model parameters for the classifiers, we run 5-fold cross validation on the development set. After the estimation of all parameters (by grid search) we train the classifier from the entire development set and apply it to the (previously unseen) test set. 

As a baseline classifier we use a linear SVM (due to its strong generalization ability and efficiency). For promising configurations, we run additional experiments with an SVM with RBF kernel and with Random Under-Sampling Boosting (RUSBoost)~\cite{seiffert2010} to investigate the influence of the classifier on the result. RUSBoost~\cite{seiffert2010} is a variant of AdaBoost~\cite{freund1996} that is optimized for classification tasks with imbalanced class priors. Classification results presented in Section \ref{sec:results} always refer to the performance obtained on the independent test set.

The SED evaluation protocol defines performance measures for both tasks~\cite{Petkos2014}. For event relevance detection the challenge defines the average of f1-scores for both classes:  $f1_{ene-avg}=(f1_{event} + f1_{non-event})/2$. For this task, all event-related images are put into one class and binary classification is performed. For event type classification the dataset is split into 9 classes (the non-event class and eight classes referring to a particular event type). The performance measure specified for the task is the average f1-score $f1_{type-avg}$ over all nine classes (e.g. $f1_{non-event}$, $f1_{concert}$, $f1_{sports}$,...)~\cite{Petkos2014}. We strictly stick to the performance measures specified by the SED evaluation protocol to assure comparability to related approaches that were also evaluated on the dataset.

We investigate the following questions in our study:

\begin{itemize}
	\item Which performance level can be achieved by contextual information only? Evaluate TF-IDF representations with different numbers of words as well as topic extraction with different numbers of topics.
	\item Do the textual features complement each other? 
	\item What is the performance level achievable by purely visual information? Extract GIST for different block sizes, as well as SIFT with sparse and dense sampling. Evaluate the performance of BoW (hard assignment) for different codebook sizes for sparse and dense SIFT. Compare BoW with VLAD codebooks of different size. Compare the performance of local features with GIST and PCA-GIST. 
	\item Do the visual features (e.g. global and local features) complement each other?
	\item Can a purely visual approach compete with an approach that exploits contextual metadata?
	\item Does the multimodal combination of textual features with visual ones facilitate classification? Which multimodal representation performs best?
	\item How sensitive are the features to their parameters?
\end{itemize}

Based on the insights gained from the performed experiments we propose a novel baseline method for social event classification and compare its results to state-of-the-art methods.

\section{Results}
\label{sec:results}

According to our experimental setup, we first present the results of purely textual processing and purely visual processing. Next, we demonstrate the capabilities of multimodal event classification by combining textual and visual information. Finally, we compare our results with that of mono- and multimodal state-of-the-art approaches.

\subsection{Purely textual classification}
\label{sec:resultsTextualFeatures}

Table~\ref{tab:textFeatures} summarizes selected (the most promising) results for purely textual analysis for event relevance detection (in terms of $f1_{event}$, $f1_{non-event}$, and $f1_{ene-avg}$) and event type classification (in terms of $f1_{type-avg}$). The row numbered with zero provides the random baseline obtained for the respective performance measures.

 \begin{table}[t]
	\centering
	\resizebox{\columnwidth}{!}{

		\begin{tabular}{c}
			
				\includegraphics[width=1\textwidth]{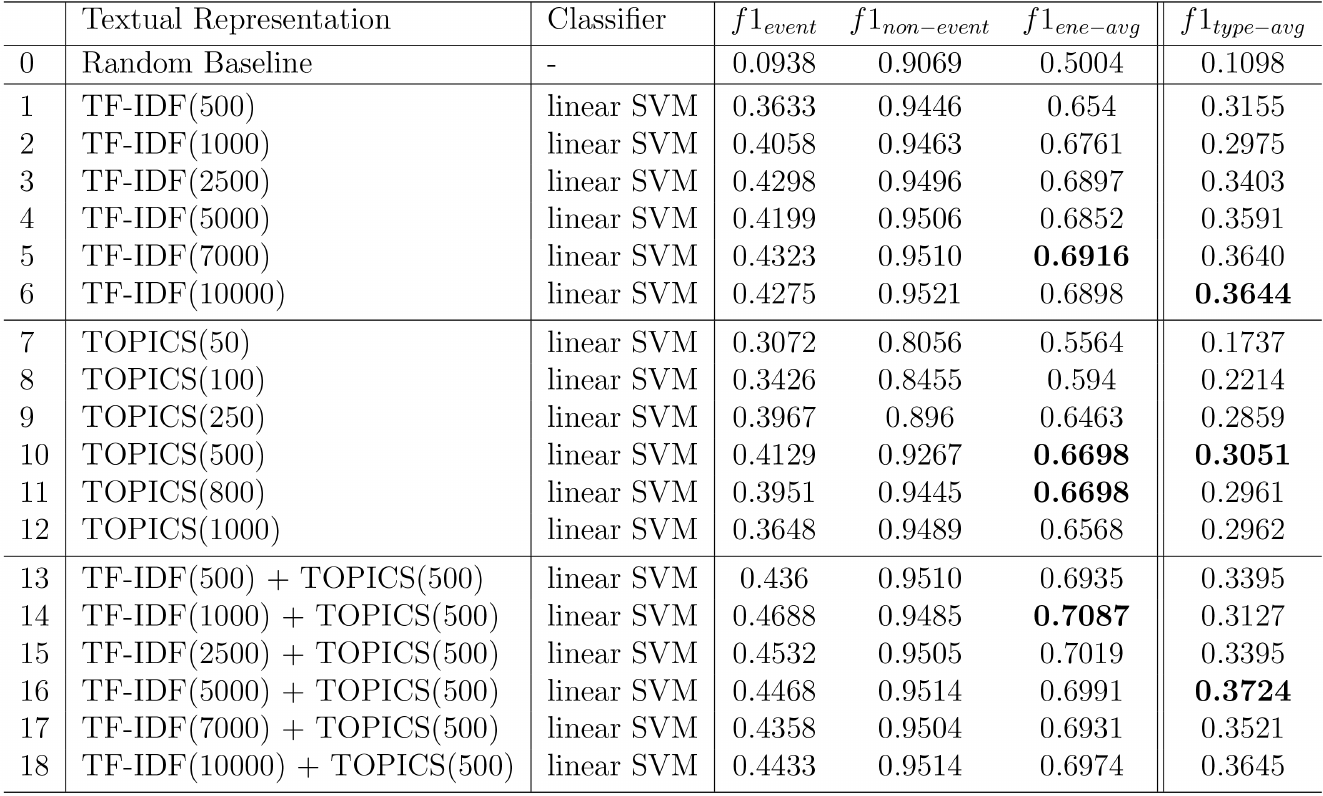}
		\end{tabular}

									}
								\caption{Textual event classification. Results for event relevance detection (columns 4-6) and event type classification (column 7). Numbers in brackets provide the dimension of TF-IDF vectors and the number of topics, respectively. The best $f1_{ene-avg}$ and $f1_{type-avg}$ scores for each representation are highlighted bold. Additional measures for each experiment are available online as supplementary material.}
								 \label{tab:textFeatures}%
							 \end{table}%

\subsubsection{Event relevance detection}

The results for TF-IDF in rows 1-6 in Table~\ref{tab:textFeatures} show a high f1 for the classification of non-event-related images ($f1_{non-event}$) above 0.94 for all evaluated TF-IDF dimensions (from 500 to 10000). The classification of event-related images yields f1 score of only 0.36-0.43. The reason for this differing behavior is the asymmetry in the dataset. The dataset contains only 6358 event-related images while the non-event class comprises 50807 images. As a consequence, the classifier is dominated by the large number of non-event images. While the number of misclassified images is similar for both classes (1341 vs. 1700), the number has much stronger influence on the f1 score of the (smaller) event-related class. We provide the random baseline for each performance measure in row 0 of Table~\ref{tab:textFeatures} to facilitate performance assessments. The random baseline for $f1_{non-event}$ is already 0.91 due to the predominance of this class. For event-related images, the baseline is significantly lower with only 0.09. Thus, the improvement from 0.09 to 0.43 by TF-IDF for event-related images represents a strong improvement over the random baseline.   

Column 6 of Table~\ref{tab:textFeatures} provides the averaged f1 score over both classes which indicates the overall classification performance. The performance increases with an increasing number of terms (from 0.65 to 0.69) which is clearly above the random baseline of 0.5. The best performance (f1 of 0.69) is obtained with TF-IDF with 7000 terms. Experiments with classifiers other than linear SVM (non-linear SVM and K-NN, not in Table~\ref{tab:textFeatures}) show that linear SVM yields the highest classification rate and is computationally most efficient.

The topic-based representation (rows 7-12 in Table~\ref{tab:textFeatures}) is slightly outperformed by TF-IDF vectors. Performance improves with increasing number of topics but the level of TF-IDF cannot be reached. Topic modeling is not always able to extract meaningful topics especially for the non-event images. A closer look at the data reveals that the tags provided for non-event-related images are often unrelated to the image or misleading. The metadata contains for example the keywords ``sport" and ``basketball" although the image has no relation to sports and just shows two children sitting on a couch. Adding more topics does not affect the averaged f1 score. We employ 500 latent topics in subsequent experiments.

Finally, we combine both features by early fusion (rows 13-18 in Table~\ref{tab:textFeatures}). The combination yields a slight improvement of overall performance to an average f1 score of 0.71. Results for the different dimensions of TF-IDF (in combination with latent topics) vary only slightly which shows that the sensitivity to this parameter is low. Higher dimensions do not necessarily lead to a higher performance. We do not observe improvements with other classifiers (e.g. non-linear SVM).

\subsubsection{Event Type Classification}

Table~\ref{tab:textFeatures} in the previous Section provides the results for event type classification in terms of average f1 score ($f1_{type-avg}$ in column 7) over all nine event classes. TF-IDF outperforms latent topics. For TF-IDF a higher dimension is beneficial, for topics a number of 500 yields the best tradeoff between performance and feature dimension. 

We observe a strong variance in performance across the different event classes. For TF-IDF the best performance is obtained for the ``protest" class. The class ``other" yields the lowest performance. This class lacks a consistent event type and thus cannot be modeled accurately. Latent topics yield similar performance than TF-IDF for the classes ``concert", ``protest", and ``conference". For all other classes f1 scores are lower. Latent topics are not able to model underrepresented event types accurately because the number of examples per class is too low to derive meaningful topics. This is for example the case for the ``fashion" class which exhibits only 36 training images. Detailed performance measures for all event classes are available in the online annex.

The combination of TF-IDF and latent topics only marginally increases the performance (+0.8\%). We do not observe stronger synergy effects between the two features in our experiments (the same is observed with other classifiers). As a baseline for further multimodal experiments with visual information (in Section~\ref{subsec:multimodalResults}) we employ TF-IDF as representation for the textual information. 

	
	\subsection{Purely visual classification}
	\label{sec:resultsVisualFeatures}
	
Similarly to the textual modality, we investigate the potentials of the visual modality by applying different visual representations (and combinations) for both investigated tasks. Table~\ref{tab:visualFeatures} presents the corresponding results.

							 \begin{table}[t]
								\centering
								\resizebox{\columnwidth}{!}{
								\begin{tabular}{c}
										
										\includegraphics[width=1\textwidth]{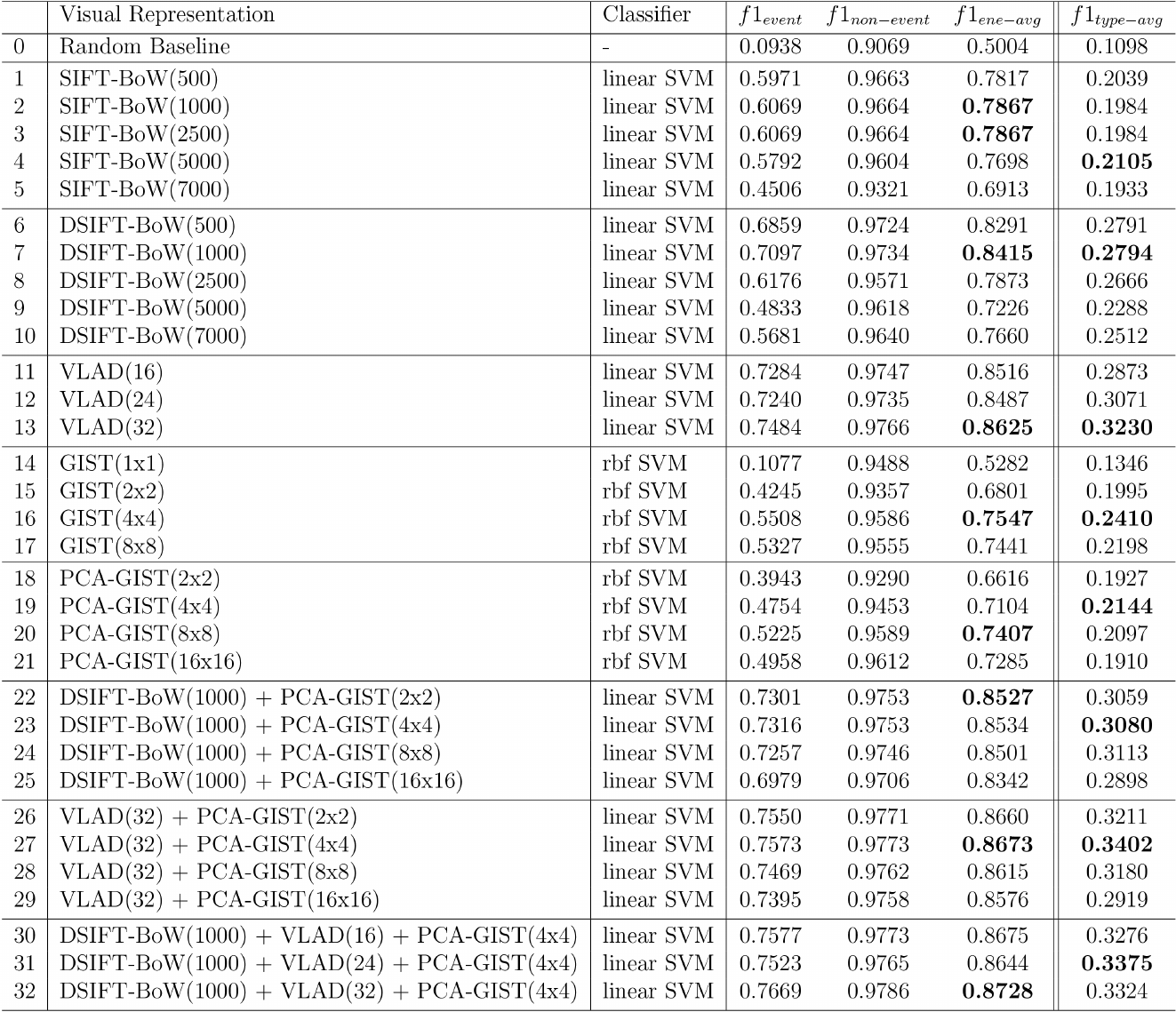}
								\end{tabular}

									}
								\caption{Event classification using only visual information. Results event relevance detection (columns 4-6) and event type classification (column 7). Numbers in brackets provide the number of clusters for BoW and VLAD representations and the number of blocks for GIST and PCA-GIST features. Additional measures for each experiment are available online as supplementary material.}
								 \label{tab:visualFeatures}%
							 \end{table}%

\subsubsection{Event relevance detection}

We first compare the classification results for BoW generated from sparse (SIFT-BoW) and dense SIFT points (DSIFT-BoW). Sparse BoW (rows 1-5 in Table~\ref{tab:visualFeatures}) yields a maximum average f1 of 0.79 for event relevance detection. This is an improvement of +7.8\% compared to the best textual approach from Section~\ref{sec:resultsTextualFeatures}. Dense BoW (rows 6-10 in Table~\ref{tab:visualFeatures}) further improves performance to an average f1 of 0.84. Dense SIFT captures more information from the images due to its better spatial coverage and thus clearly outperforms sparse SIFT in this task. The recognition of non-event related images can be accomplished nearly completely with DSIFT-BoW (best f1 score for non-events 0.97). This is an improvement of +2.2\% compared to the best textual approach. For the event class performance improves strongly by visual analysis compared to the textual approach (+24,09\%). These results demonstrate that visual information is crucial for the task.

Next, we investigate the performance of VLAD features (rows 11-13 in Table~\ref{tab:visualFeatures}). While the f1 for non-events increases only slightly to 0.98, the f1 for events increases by +3.87\% to 0.75 resulting in an overall (average) f1 of 0.86.  VLAD outperforms SIFT-BoW and DSIFT-BoW and all approaches based on purely textual information. The representations evaluated so far yield the best results in combination with a linear SVM.

Next, we evaluate the global image representations GIST and PCA-GIST. GIST requires a more flexible kernel such as RBF to achieve competitive results (rows 14-18 in Table~\ref{tab:visualFeatures}) which is however at the cost of processing time. For GIST with more than 8x8 blocks classification did not terminate. The overall performance of GIST features is lower than that of the local features (BoW and VLAD) with a maximum f1 score of 0.75 with 4x4 blocks. 

PCA-GIST has 6-times less components than GIST. They enable much faster classification and yield a similar performance level than GIST (rows 18-21 in Table~\ref{tab:visualFeatures}). Again, a non-linear kernel outperforms the linear one. 

Next we evaluate different combinations of local and global features: DSIFT-BoW + PCA-GIST (rows 22-25 Table~\ref{tab:visualFeatures}), VLAD + PCA-GIST (rows 26-29 Table~\ref{tab:visualFeatures}) and the combination of all three features: DSIFT-BoW, VLAD, PCA-GIST (rows 30-32 Table~\ref{tab:visualFeatures}). We employ PCA-GIST instead of GIST because of their computational efficiency and similar performance. 

All combinations of global and local features marginally improve results. The best combination is that of all three features which improves performance by +1.03\% over the best individual feature (VLAD) and yields an overall performance of 0.87. The other combinations show that addition of global features adds only little benefit to event relevance detection. We assume that the heterogeneity of the image compositions for event and non-event images is too high to derive useful information from GIST. 

Local features (especially VLAD) perform well and clearly outperform textual features. The best result obtained by a purely visual approach is 0.87 while the best textual approach yields only 0.71. A reason for this behavior might be the different visual appearance of the images in the two classes. While the event-related images frequently show places, stages, halls, and play fields, the non-event related images often capture portraits of people and images of products, see also Figure~\ref{fig:imagedata} in Section~\ref{sec:Intro} for examples.

\subsubsection{Event type classification}

\begin{figure}[t]%
			\centering
			\includegraphics[width=0.7\linewidth]{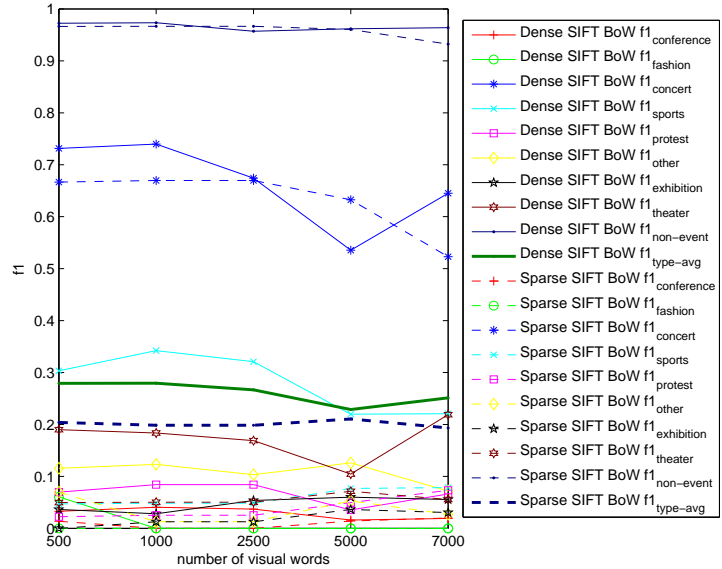}
		\caption{Performance of sparse and dense SIFT BoW for event type classification for all event types and codebook sizes.}
		\label{fig:siftFeatures_type}
	\end{figure}

The results for event type classification by purely visual information are listed in Table~\ref{tab:visualFeatures} column 7. Dense BoW (with 1000 terms) yields an average f1 of 0.28 and outperforms sparse BoW with +6.89\%. Figure~\ref{fig:siftFeatures_type} shows the performance of the sparse and dense SIFT BoW features for all event types. The concert class can be discriminated best (aside from the non-event class). For the other event classes f1 scores of dense BoW are below 0.35 and for sparse BoW even below 0.21. The lowest score is obtained for the ``fashion" class which is underrepresented in the dataset. The sensitivity of the representations to the codebook size is low.

VLAD outperforms sparse and dense BoW representations (Table~\ref{tab:visualFeatures} rows 11-13). For 32 codewords an overall f1 score of 0.32 is obtained and all classes (except the fashion class) yield f1 scores larger than 0.1.  GIST and PCA-GIST achieve weaker results than the local image representations. The classes ``concert" and ``sports" are best represented. There is, however, no event type for which global features outperform local ones. 

The combination of visual features yields a slight improvement of performance. DSIFT-BoW combined with PCA-GIST improves by +2.85\% (Table~\ref{tab:visualFeatures} rows 22-25). VLAD combined with PCA-GIST improves by +1.72\% (Table~\ref{tab:visualFeatures} rows 26-29) which is the highest result obtained by purely visual processing. The combination of all three features (Table~\ref{tab:visualFeatures} rows 30-32) does not further improve performance which may be attributed to the redundancy of the VLAD and BoW features.   

\begin{figure}[t]%
			\centering
				\includegraphics[width=0.7\linewidth]{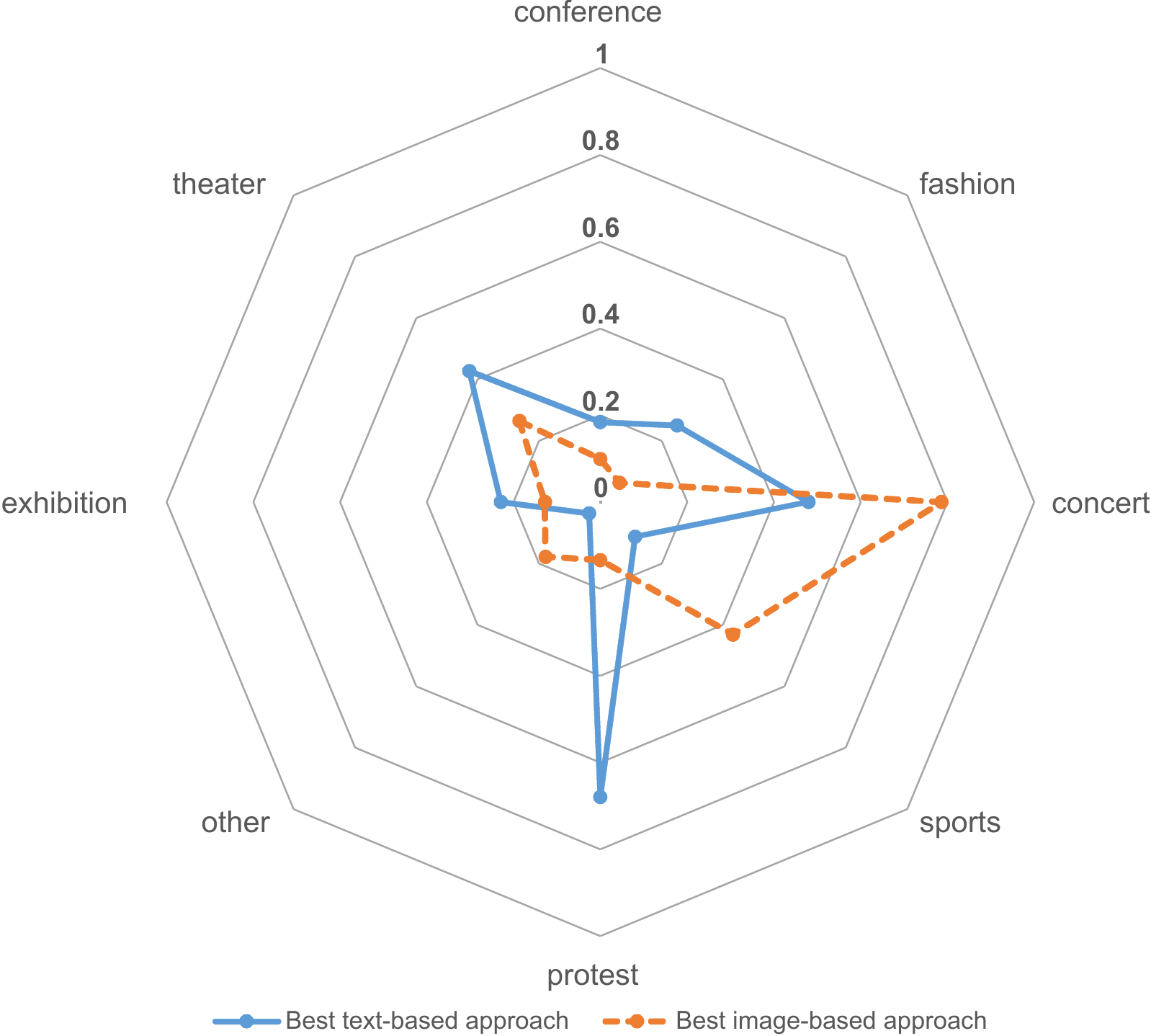}
			\caption{Textual vs. visual event type classification. F1 scores for different event types for the best visual and textual approaches. Both approaches complement well each other.}
			\label{fig:textVsvisualEventTypeClass}
	\end{figure}

In comparison to textual features we observe that visual features cannot achieve the same performance level for event type classification. The best textual approach (TF-IDF + TOPICS) with an f1 of 0.37 is still 3.22\% better than the best visual approach (VLAD + PCA-GIST). However, we observe complementary behavior between textual features and visual features over different event types. For three classes visual features strongly outperform textual features in f1 score: ``concert": 0.78 vs. 0.48, ``sports": 0.43 vs. 0.11, and ``other": 0.18 vs. 0.04, see Figure~\ref{fig:textVsvisualEventTypeClass}. The difference in both polylines illustrates well the complementary character of the textual and visual approaches. The insights gained in the experiments so far give rise to the assumption that textual and visual information are well-suited for combination in a multimodal approach.

	\subsection{Multimodal classification}
	\label{subsec:multimodalResults}

Table~\ref{tab:multimodalApproach} shows results obtained by different multimodal representations. The combination of visual and textual information is performed by early fusion (concatenation). A comparison to late fusion strategies is presented in Section~\ref{subsec:fusionStrategies}.

					\begin{table}[t]
								\centering
								\resizebox{\columnwidth}{!}{
								\begin{tabular}{c}
									
										\includegraphics[width=1\textwidth]{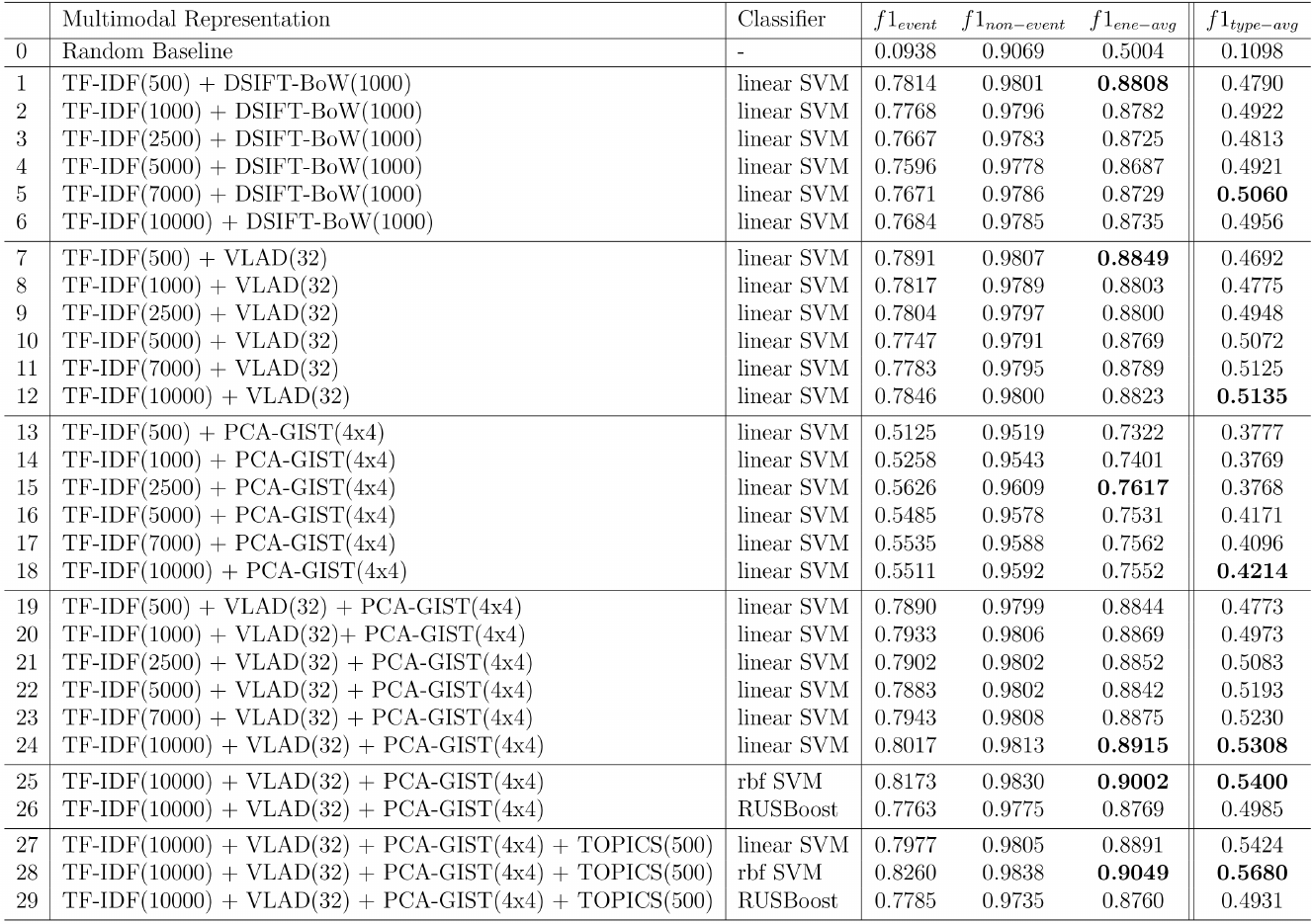}
								\end{tabular}

									}
								\caption{Event classification using multimodal information. Results for event relevance detection (columns 4-6) and event type classification (column 7). The multimodal representations outperform the purely textual and purely visual representations for both investigated tasks. Additional measures for each experiment are available online as supplementary material.}
								 \label{tab:multimodalApproach}%
							 \end{table}%

\subsubsection{Event relevance detection}
\label{sec:ENEMultimodal}
The best result so far has been obtained by combining global and local visual features. The combination of visual information with contextual information further improves results. We combine TF-IDF with DSIFT-BoW (rows 1-6 in Table~\ref{tab:multimodalApproach}), TF-IDF with VLAD (rows 7-12) and TF-IDF with PCA-GIST (rows 13-18). In all three experiments, the multimodal representation improves performance combined to the respective individual features (+3.93\% for TF-IDF+DSIFT-BoW, +2,24\% for TF-IDF+VLAD, and +2.1\% for TF-IDF+PCA-GIST).

Next, we add global \emph{and} local visual information to the textual representation (TF-IDF+VLAD+PCA-GIST, rows 19-24 in Table~\ref{tab:multimodalApproach}). This combination further improves the results to an average f1 of 0.89 with an $f1_{event}$ of 0.80 and an $f1_{non-event}$ of 0.98. We evaluate this combination with different classifiers to see how well the linear SVM models the data. An SVM with RBF kernel (row 25) further improves classification performance to 0.90 while RUSBoost (row 26) yields a slightly weaker performance of 0.88. These results confirm that the linear SVM provides a good performance tradeoff, especially when we consider the significantly lower run-time. 

In a final experiment, we additionally add latent LDA topics to our multimodal representation. The linear SVM and RUSBoost (rows 27 and 29 in Table~\ref{tab:multimodalApproach}) cannot take advantage of the additional information. The SVM with RBF kernel, however, further improves results and yields an average f1 of 0.905 (the peak performance obtained in our experiments). By combining both modalities we obtain an improvement of +3.21\% compared to the best monomodal result and strongly outperform the baselines for both event and non-event classes.

\subsubsection{Event type classification}

The experiments on purely textual and purely visual classification in Sections~\ref{sec:resultsTextualFeatures} and~\ref{sec:resultsVisualFeatures} indicate a strong complementary behavior of both modalities for event type classification. The results in Table~\ref{tab:multimodalApproach}, column 7 confirm this assumption. The best average f1 obtained from purely textual information is 0.37 and from purely visual information 0.34. The combination of TF-IDF with local image representations (DSIFT-BoW and VLAD) increases performance up to 0.51 which corresponds to a gain of +14.91\%. Adding global features (PCA-GIST) to TF-IDF yields an improvement of +5.7\%.

The combination of textual information with both global and local image representations further improves results to 0.53. Again we evaluate the multimodal representation with RUSBoost and RBF SVM (rows 25 and 26, Table~\ref{tab:multimodalApproach}) and observe an improvement through the RBF kernel to 0.54. The addition of the topic-based representation (row 28) further improves results to 0.568 with RBF SVM. The results confirm that the selected features capture relevant and complementary information for event type classification. 

\begin{figure}[t]%
			\centering
				\includegraphics[width=0.70\linewidth]{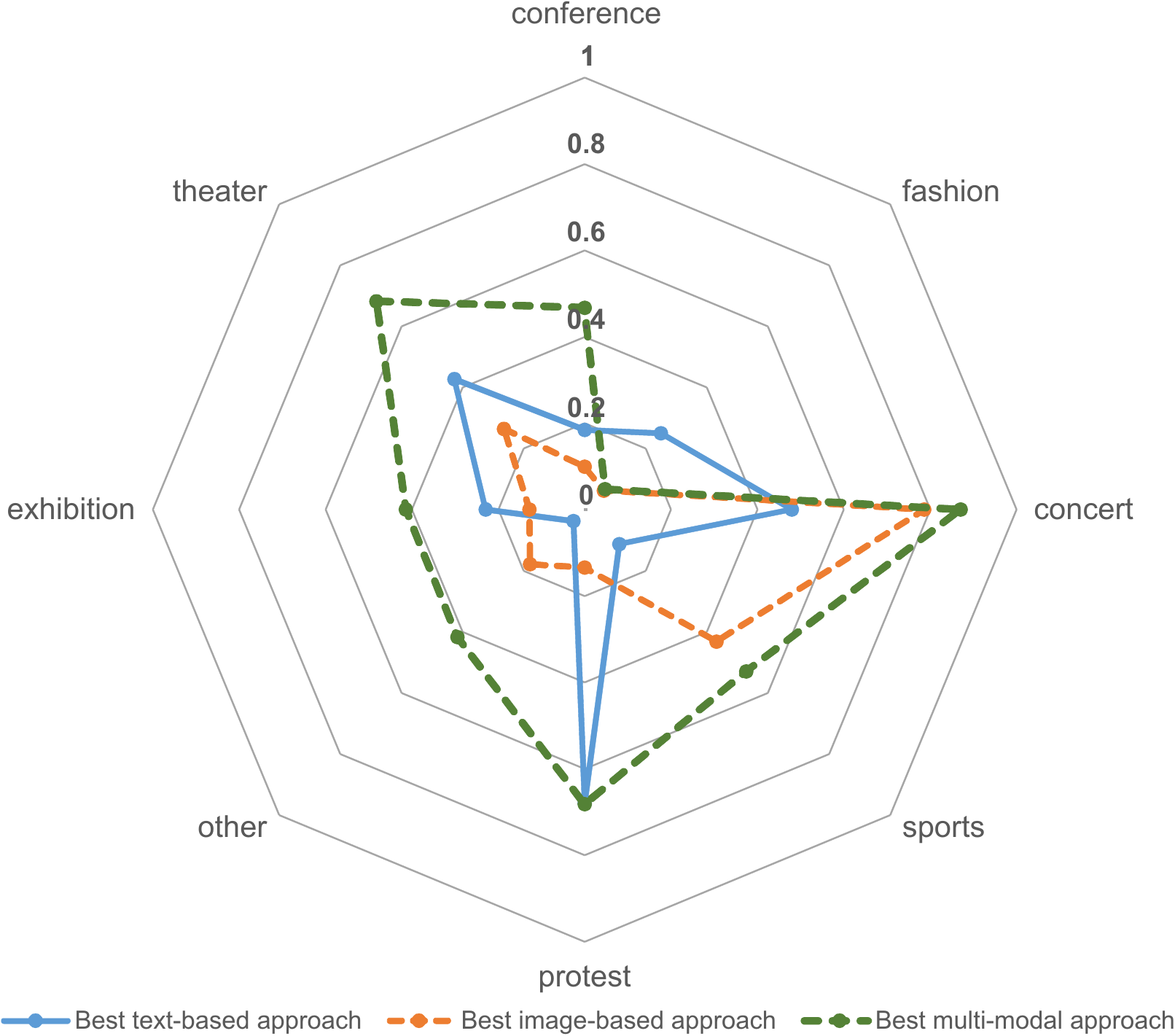}
			\caption{Monomodal vs. multimodal event type classification. F1 scores for different event types for the best monomodal approaches and the best multimodal approach. The multimodal approach outperforms the purely textual and visual ones for all classes except the ``fashion" class.}
			\label{fig:monoVsMultimodalEventTypeClass}
			\end{figure}

Figure~\ref{fig:monoVsMultimodalEventTypeClass} illustrates the difference between the best multimodal approach and the best monomodal approaches. For all classes except the underrepresented ``fashion" class the results are improved by multimodal processing. The beneficial effect is demonstrated well by the example of the ``concert" and ``protest" classes. The class ``concert" is represented well by the visual approach, but not by the textual approach. The opposite is the case for the ``protest" class where the purely visual approach fails and the textual approach yields high performance. The multimodal approach achieves high performance for both classes which clearly shows that the two modalities attain synergy.

\subsection{Fusion strategies }
\label{subsec:fusionStrategies}

From all experiments performed so far, we select the most promising configurations for purely textual\footnote{T: TF-IDF(5000)+TOPICS(500) with linear SVM}, visual\footnote{V: VLAD(32)+PCA-GIST(4x4) with linear SVM}, and multimodal\footnote{T+V:TF-IDF(10000)+VLAD(32)+PCA-GIST(4x4)+TOPICS(500) with RBF SVM} classification and apply different feature fusion strategies. Additionally to early fusion (performed so far), we investigate the two late fusion strategies described in Section \ref{subsec:classification}.	In Table~\ref{tab:fusionStrategies} we provide the f1-scores as in previous sections as well as recall ($R_{event}$, $R_{non-event}$) and precision ($P_{event}$, $P_{non-event}$) for event relevance detection and the f1-scores for each individual event type. We list the performance on the test set (``test"), the development set (averaged over all cross-validation runs, $\mu(dev)$) and the respective standard deviations $\sigma(dev)$ to evaluate the robustness of the approach to different training partitions. The main findings from our experiments are the following:

\begin{itemize}
 \item Early fusion in most cases outperforms late fusion (especially for event type classification). We do not observe a significant improvements with late fusion on the test set. Hierarchical late fusion outperforms in most cases additive late fusion. A reason for this might be the additional abstraction introduced by the top-level classifier in hierarchical late fusion. The stronger performance of early fusion indicates that the higher-dimensional input space (due to concatenation of the features) facilitates classification and that the SVM is able to exploit this high-dimensional information.
 \item In all experiments except for one (event relevance detection with additive late fusion) the multimodal approach outperforms purely visual and textual ones.
 \item The standard deviations are in most cases small ($<$0.02), especially in event relevance detection. For event type classification we observe higher standard deviations, e.g., for the three smallest classes in the development set: ``fashion", ``conference", and ``protest". We assume that the lack of training data makes the classes difficult to model (especially when cross-validation further reduces the amount of training data). For classes with high cardinality (e.g. ``concert") the standard deviations are consistently low ($\leq$0.02).
\item The linear SVMs employed in early fusion and additive late fusion generalize well from the training data. The multimodal approaches achieve even higher performance on the test set than on the development set. The combination of numerous features from different modalities seems to improve robustness. A different trend can be observed for hierarchical late fusion. Here, we employ an SVM with RBF kernel (as it outperformed the linear kernel, see Table~\ref{tab:multimodalApproach}). The RBF kernel increases training performance significantly. The classifier can, however, not achieve a similar performance on the test set which indicates overfitting during training. 
\end{itemize}

Additional results from the performed experiments including confusion matrices for experiments with different fusion strategies are available online (as supplementary material) in the electronic annex.

							 \begin{table}[!ht]
								\centering
								\resizebox{!}{.3\paperheight}{ 
			
								\begin{tabular}{c}
														\includegraphics[width=1\textwidth]{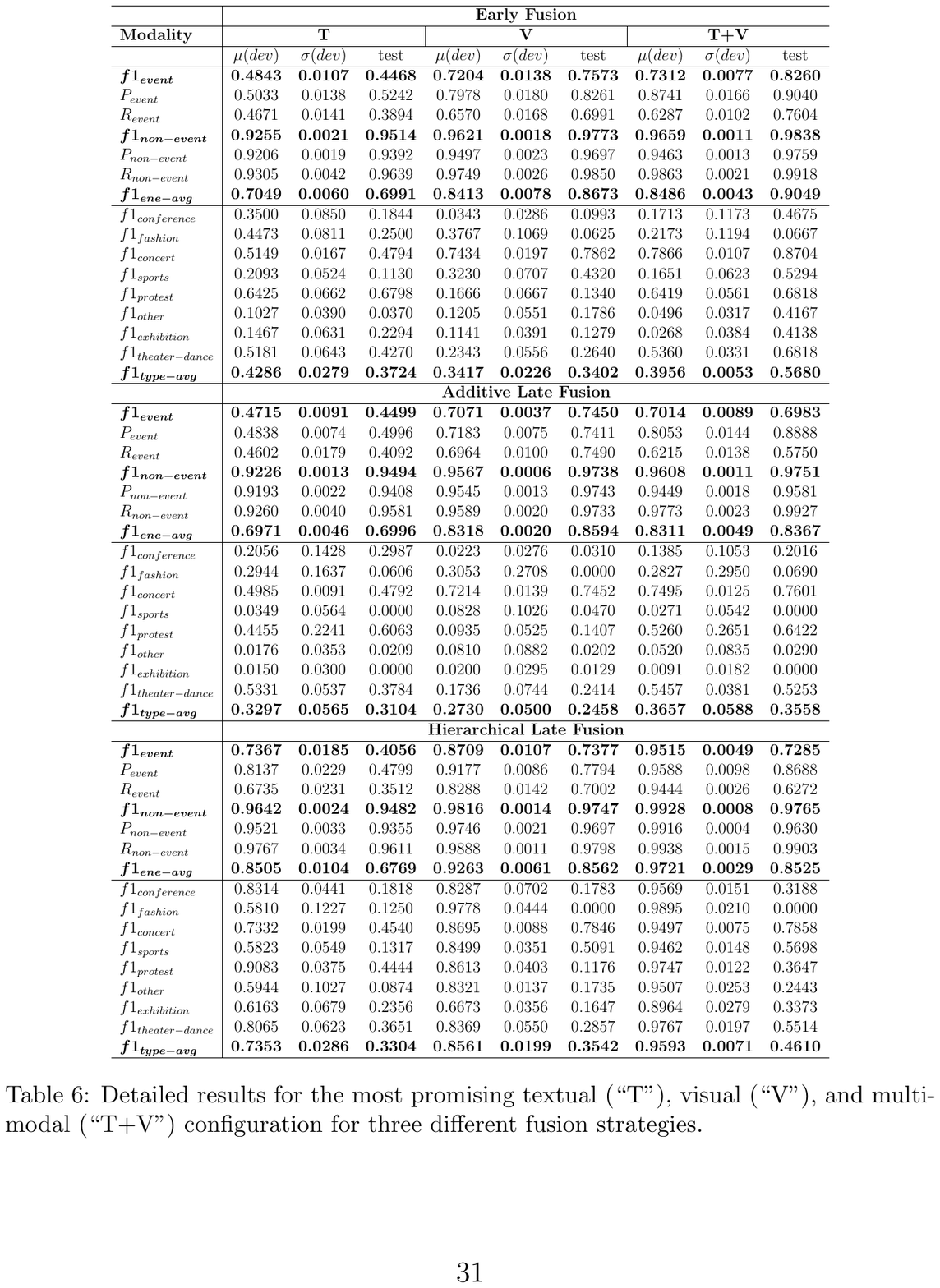}
								\end{tabular}

									}
								\caption{Detailed results for the most promising textual (``T"), visual (``V"), and multimodal (``T+V") configuration for three different fusion strategies.}
								 \label{tab:fusionStrategies}%
							 \end{table}%

	\subsection{Comparison to the state-or-the-art}
	\label{comparisonSOTA}
	
	To conclude our experiments we compare our results with that of comparable state-of-the-art methods which have been developed or evaluated in the course of the MediaEval SED challenge, see Table~\ref{tab:results}. Results are taken from the original papers and from~\cite{Petkos2014}. Additional results were kindly provided by the SED organizers. Measures that could not be retrieved were left empty. 

\begin{table}[t]%
\resizebox{\columnwidth}{!}{
			\begin{tabular}{c}
					
					\includegraphics[width=1\textwidth]{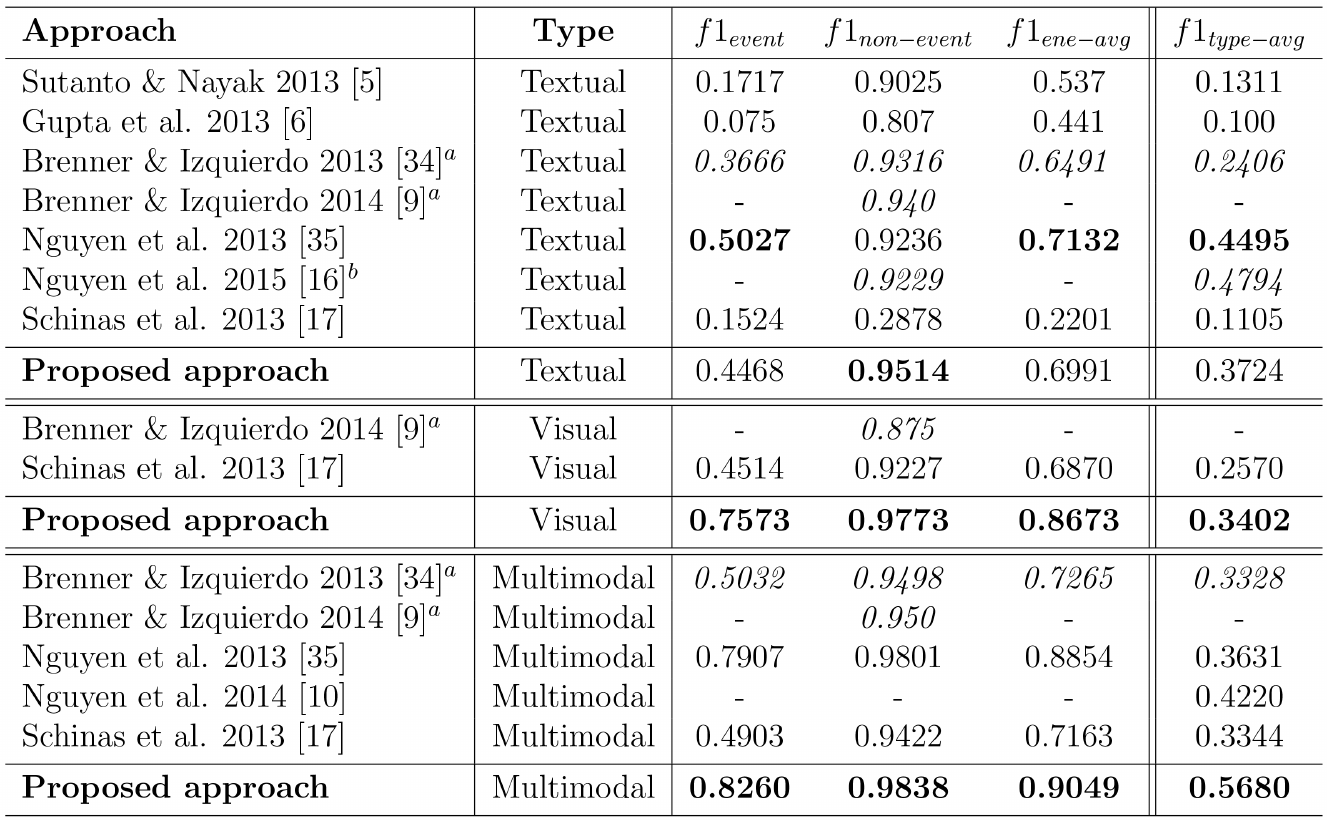}
			\end{tabular}

}

\caption{Results of textual, visual, and multimodal state-of-the-art methods and the proposed approach. The best results of directly comparable methods are bold. Results set italic are not directly comparable due to $^a$ and $^b$.
\newline  $^{a)}$ The authors perform cross-validation across both, the development and test set. We perform cross-validation only on the development set. 
\newline $^{b)}$ The authors evaluate only on the development set.}
\label{tab:results}
\end{table}

Our purely textual method outperforms all other text-based approaches except that of Nguyen et al.~\cite{Nguyen-SED2013} (especially for event type classification). In contrast to our approach, Nguyen et al.~\cite{Nguyen-SED2013} include data from external sources (from a large ontology). We assume that the additional external information explains the higher performance. Note that the approach of~\cite{Nguyen2015} also seems to outperform our approach (especially for event type classification). This is however questionable, since the authors state that they used only the SED development set for evaluation which contains only half of the data.

Only ~\cite{brenner2014} and~\cite{CERTH-SED2013} report results on purely visual classification. The results are weaker that that of our purely visual approach. A reason for the higher performance of our approach is the combination of local and global image information in one representation, whereas~\cite{brenner2014} and~\cite{CERTH-SED2013} employ either local or global information.

The best results for event classification are obtained by multimodal approaches. Our multimodal approach outperforms all other approaches. For event relevance detection we improve the state-of-the-art of 0.885 of~\cite{Nguyen-SED2013} to 0.905. The best result for event type classification by a related approach is an average f1 of 0.422. Our approach surpasses  this result by +14.6\% (f1 of 0.568). An interesting observation from this result is that the more advanced visual features in our approach easily compensate the advancements obtained by the more complex textual processing of~\cite{Nguyen2014}. This confirms the strong importance of visual information for event type classification.

\section{Conclusions}
\label{sec:conclusions}

Social event classification is an important task for the indexing and retrieval of event-related content shared on social media platforms. In this paper we presented a comprehensive study on social event classification. We investigated the capabilities of textual and visual representations and studied the multimodal nature of the task. While textual information is more important for event type classification, visual information shows to be of higher importance for event relevance detection. The combination of textual and visual information strongly improves both tasks. The obtained results on the publicly available SED benchmark dataset show that our approach outperforms state-of-the-art approaches and thus represents a novel baseline for future research. 




\newpage 
		
\section*{References}

  \bibliographystyle{elsarticle-num} 
 \bibliography{mybibfile}

\newpage 

%
%
%
\end{document}